\documentclass[letterpaper]{article} 
\usepackage[]{aaai2026}  
\usepackage{comment}
\usepackage{times}  
\usepackage{helvet}  
\usepackage{courier}  
\usepackage[hyphens]{url}  
\usepackage{graphicx} 
\usepackage{amsmath}
\usepackage{amssymb}
\usepackage{natbib}  
\usepackage{caption} 
\usepackage{algorithm}
\usepackage{algorithmic}

\usepackage{booktabs} 
\usepackage{multirow} 
\usepackage{makecell} 
\usepackage{empheq}   
\usepackage{xcolor}   
\usepackage{amsthm}   
\newtheorem{theorem}{Theorem} 

\usepackage{newfloat}
\usepackage{listings}
\DeclareCaptionStyle{ruled}{labelfont=normalfont,labelsep=colon,strut=off} 
\floatstyle{ruled}
\newfloat{listing}{tb}{lst}{}
\floatname{listing}{Listing}

\title{GMoPE: A Prompt-Expert Mixture Framework for Graph Foundation Models}
\author{
    Zhibin Wang\textsuperscript{\rm 1}\thanks{Equal contribution.}, 
    Zhixing Zhang\textsuperscript{\rm 1},
     Shuqi Wang\textsuperscript{\rm 1},
    Xuanting Xie,\textsuperscript{\rm 1}
    zhao Kang\textsuperscript{\rm 1}\footnote{Corresponding author.}
}
\affiliations{
 \textsuperscript{\rm 1}School of Computer Science and Engineering, University of Electronic Science and Technology of China \\

}

\usepackage{bibentry}

\begin{document}
\makeatletter
\def\copyright@text{}
\makeatother

\maketitle

\begin{abstract}
Graph Neural Networks (GNNs) have demonstrated impressive performance on task-specific benchmarks, yet their ability to generalize across diverse domains and tasks remains limited. Existing approaches often struggle with negative transfer, scalability issues, and high adaptation costs. To address these challenges, we propose \textbf{GMoPE} (Graph Mixture of Prompt-Experts), a novel framework that seamlessly integrates the Mixture-of-Experts (MoE) architecture with prompt-based learning for graphs. \textbf{GMoPE} leverages expert-specific prompt vectors and structure-aware MoE routing to enable each expert to specialize in distinct subdomains and dynamically contribute to predictions. To promote diversity and prevent expert collapse, we introduce a soft orthogonality constraint across prompt vectors, encouraging expert specialization and facilitating a more balanced expert utilization. Additionally, we adopt a prompt-only fine-tuning strategy that significantly reduces spatiotemporal complexity during transfer. We validate \textbf{GMoPE} through extensive experiments under various pretraining strategies and multiple downstream tasks. Results show that \textbf{GMoPE} consistently outperforms state-of-the-art baselines and achieves performance comparable to full parameter fine-tuning—while requiring only a fraction of the adaptation overhead. Our work provides a principled and scalable framework for advancing generalizable and efficient graph foundation models.

\end{abstract}

\begin{links}
\end{links}

\section{Introduction}
Graph-structured data has become one of the most prevalent and versatile data forms, with widespread applications across citation networks~\cite{a2}, recommender systems~\cite{a1}, and biochemical analysis~\cite{a4}. Motivated by the success of large language models (LLMs) ~\cite{foundationmodel}, there is growing momentum to develop foundation models tailored for graph data. Achieving both strong generalization performance and training efficiency in this context is widely regarded as a key step—and potential milestone—toward realizing truly general-purpose graph foundation models~\cite{Samgpt,mutikonwledgegraph}.

However, enabling effective cross-domain pretraining and transfer learning on graphs remains highly challenging, primarily due to two fundamental obstacles. First, the intrinsic complexity of graph data—including irregular topologies, sparsity, and intricate dependency structures—leads to significant spatiotemporal overhead in model training~\cite{GNNsurvey}, adaptation, and transfer. Second, conventional graph neural network (GNN) architectures are often tightly coupled with specific structural assumptions (e.g., homophily or fixed node alignment), which notably limits their generalizability across diverse domains and tasks, particularly under distribution shifts~\cite{homophily,coordinatealignment,topoAlignment}.

The Mixture of Experts (MoE) architecture—well established in NLP for improving scalability and transferability~\cite{Moebase,switchtransformers}—has recently gained attention in the context of graph learning. Prior efforts to integrate MoE into GNN have primarily focused on enhancing message-passing mechanisms~\cite{GraphDIVE,GMoE,multilayermoe}, with relatively limited exploration of its potential for cross-domain pretraining and transfer learning. 

Concurrently, prompt learning \cite{promptbase} has been introduced into graph learning as a means of enabling efficient adaptation to downstream tasks. This approach aims to match—or even surpass—the performance of full parameter fine-tuning by freezing most model parameters and optimizing only lightweight, task-specific prompt components\cite{promptpower}. Pioneering work such as GPF~\cite{GPF} has demonstrated promising empirical results. However, designing effective prompts for cross-domain scenarios remains an open challenge, as it introduces new layers of complexity related to generalization, alignment, and robustness.

\begin{figure}[t]
\centering
\includegraphics[width=0.8\columnwidth]{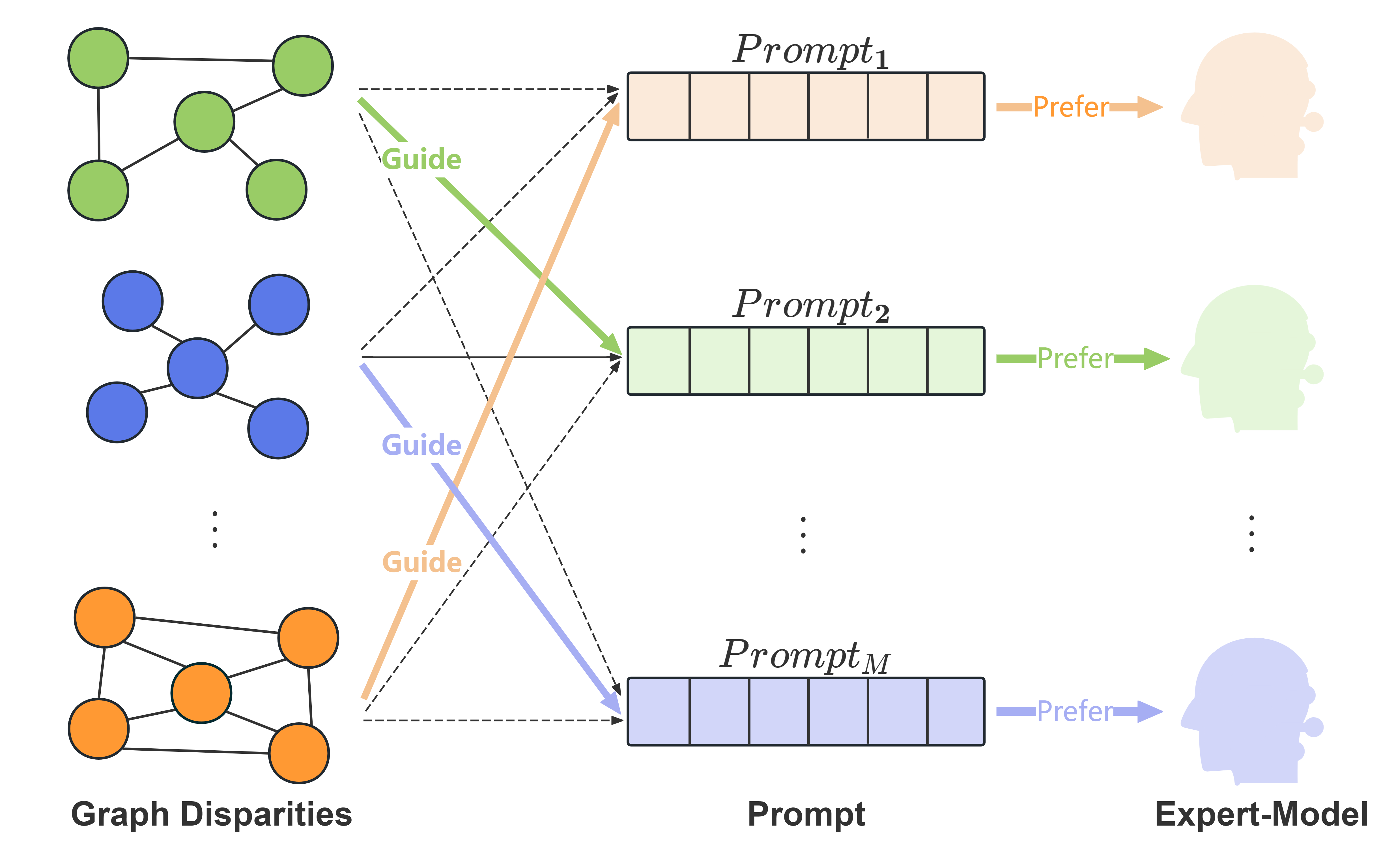} 
\caption{Guide experts with prompts}
\label{fig1}
\end{figure}

To address above challenges, this work introduces Graph Mixture of Prompt-Experts (GMoPE), a unified framework that integrates graph prompting with the MoE architecture to advance graph foundation models. As illustrated in Fig.~\ref{fig1}, GMoPE uses expert-specific, learnable prompts to guide experts in adapting to diverse graph structures, promoting specialization while maintaining scalability and transferability. By adaptively routing inputs to prompt-conditioned experts and fine-tuning only lightweight prompt parameters during downstream tasks, GMoPE achieves efficient, modular, and robust graph representation learning. Additionally, a soft orthogonality loss is introduced to encourage diversity among expert representations, effectively mitigating expert collapse and further enhancing model robustness.\\
Based on the above insights, our contributions focus on addressing the following three fundamental challenges:
\begin{itemize}
    \item \textbf{C1:}  How can we design a scalable and generalizable MoE-based framework capable of effective pretraining on cross-domain graph datasets, while maintaining strong downstream performance with minimal adaptation?
    \item \textbf{C2:} How can we effectively mitigate the expert collapse issue in graph-specialized MoE architectures to ensure structural diversity and expert specialization?

\item \textbf{C3:} How can we reduce the spatiotemporal complexity of model transfer across diverse graph domains and tasks, without compromising performance?

\end{itemize}

\section{Related Work}
\subsection{Graph Foundation Model}
Graph foundation models (GFMs) have recently emerged as unified pretrained architectures designed to learn transferable representations across diverse graph domains and tasks~\cite{graphfoundationmodelssurvey}. Traditional GFM approaches typically pretrain GNNs using strategies such as GAE \cite{GAE}, DGI \cite{DGI}, or GraphCL \cite{GraphCL}, followed by full-parameter fine-tuning on downstream tasks. However, due to the inherent negative transferability exhibited by GNNs \cite{homophily,graphfewshot}, these methods often struggle to generalize effectively intiple domains. Moreover, the large number of parameters in GNNs makes full fine-tuning computationally expensive and prone to catastrophic forgetting.

Prompt-based fine-tuning techniques~\cite{graphpromptbase} have recently been introduced in graph learning, where only a small set of prompt vectors are updated while the core GNN parameters remain frozen. Early work utilized prompts primarily for unified multitask training and task encoding~\cite{graphpromptbase}, enabling task adaptation but lacking tailored structural design. GraphPrompt~\cite{graphprompt} integrates prompt vectors with node embeddings during downstream adaptation; however, this method does not fundamentally modify the model’s input structure or the information flow within GNN layers. GPF~\cite{GPF} provides theoretical support for prompt-based feature fusion in guiding GNN transfer, but since all tasks share a single GNN backbone and prompts only perturb input features, it lacks structural adaptability and may underperform on graphs with diverse topologies. Building on GPF, Multigprompt~\cite{multigprompt} employs a dual-prompt design (composed and open prompts) to reconstruct inputs and adapt to new tasks; yet, its focus remains primarily on task generalization rather than structural generalization.

Building on these insights, we present the first framework that integrates graph prompt learning with the MoE architecture. This novel combination effectively mitigates the limitations of traditional GNNs in multi-domain joint training and addresses the  issue of expert collapse in standard MoE designs. Moreover, our method is fully compatible with existing pretraining paradigms and supports efficient downstream adaptation via prompt-based fine-tuning, ensuring minimal computational overhead during transfer.
\begin{figure*}[t]
\centering
\includegraphics[width=0.9\textwidth]{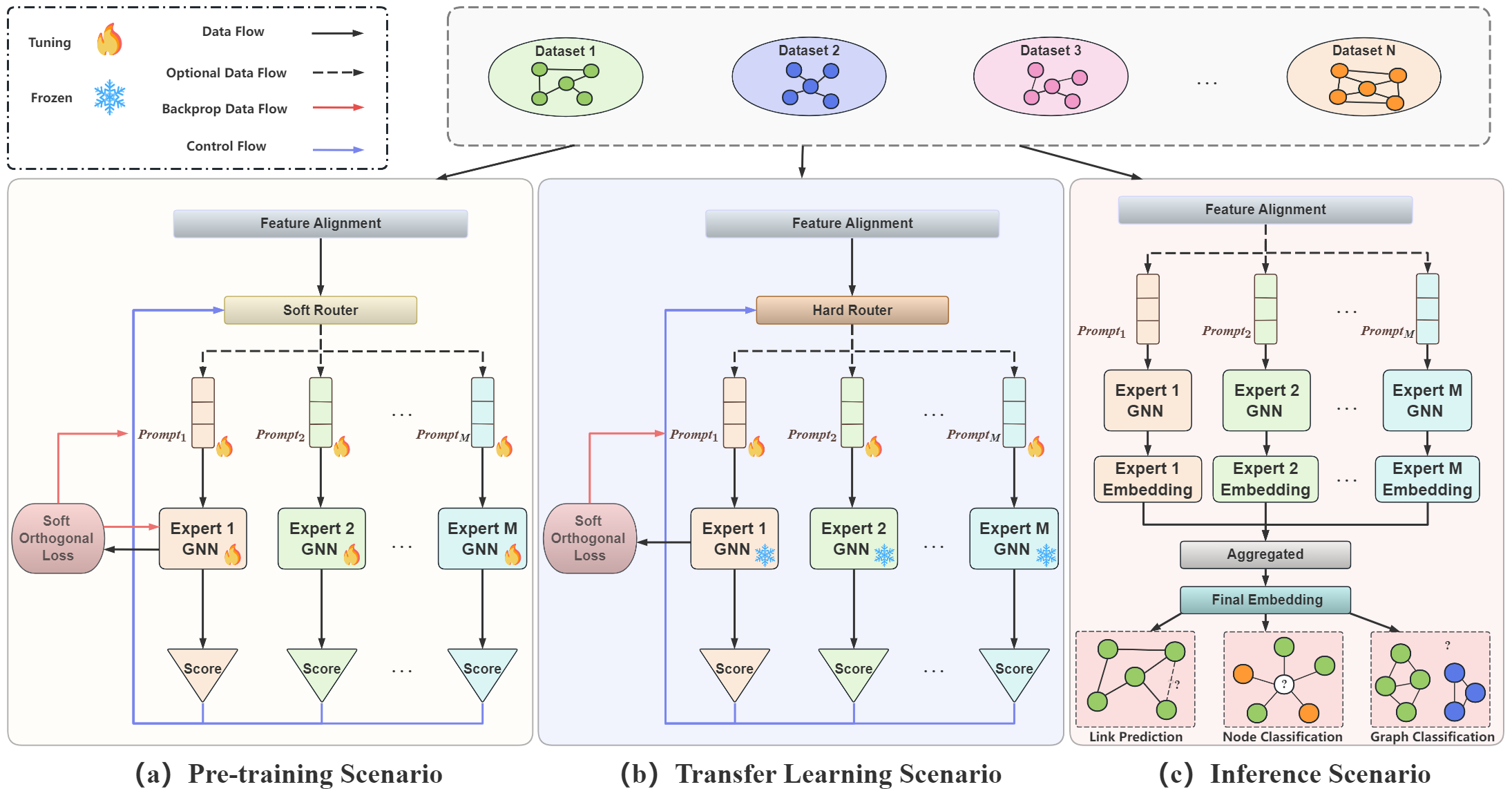} 
\caption{Overview of the proposed GMoPE framework. (a) Pre-training phase: All model parameters are optimized jointly, guided by structure-aware MoE routing and a soft orthogonality loss to encourage expert diversity.
(b) Transfer learning phase: Expert parameters are frozen, and only task-specific prompts are fine-tuned, enabling efficient and modular adaptation to new domains.
(c) Inference phase: Expert outputs are aggregated based on learned routing to produce the final graph representations for downstream tasks.}
\label{fig2}
\end{figure*}

\subsection{Mixture of Experts}
A growing body of research has explored the integration of MoE architectures into graph learning. GraphDIVE~\cite{GraphDIVE} leverages MoE to mitigate class imbalance by grouping structurally similar graphs and allowing specialized experts to focus on underrepresented classes. GMoE~\cite{GMoE} incorporates the MoE framework into the message-passing paradigm of GNN, enhancing the model’s ability to adapt to diverse graph topologies. These studies demonstrate the potential of MoE in improving representational capacity and specialization in graph-based models. MoE-ML-LP~\cite{multilayermoe} applies MoE to link prediction in multilayer networks, offering early evidence of MoE’s potential in addressing more complex and heterogeneous graph structures.
AnyGraph~\cite{AnyGraph} applies the MoE architecture to GFMs, addressing expert collapse primarily through conventional load-balancing loss functions. However, it does not fundamentally leverage the inherent, domain-specific structural characteristics of graph data. Furthermore, the complexity of its MoE design introduces significant computational overhead, which constrains its effectiveness as a lightweight and efficient foundation model for fast adaptation in graph learning.

Building on these observations, we propose an enhanced MoE architecture specifically tailored to the structural characteristics of GNNs. Our framework is designed to be broadly compatible with a wide range of GNN variants, facilitating scalable and efficient pretraining as well as transfer learning across heterogeneous graph domains.

\section{Preliminaries }
\subsection{Graph Encoder}
A graph is defined as \( \mathcal{G} = (V, \mathcal{E}, X) \), where \( V \) is the set of nodes, \( \mathcal{E} \subseteq V \times V \) represents the set of edges, and \( X \in \mathbb{R}^{|V| \times D} \) is the node feature matrix. \\
GNNs such as GCN~\cite{a2}, GAT~\cite{GAT}, GraphSAGE~\cite{GraphSAGE}, and GIN~\cite{GIN} are widely adopted as graph encoders. These models use message passing to iteratively update node representations by aggregating information from local neighborhoods. In our framework, any standard GNN can serve as an expert in the MoE system, allowing flexible integration of existing graph encoders.
\subsection{MoE Router}
An MoE router is a mechanism that assigns input-dependent weights to a set of expert models. 
Given an input \( x \), let \( \{E_1, E_2, \ldots, E_M\} \) denote the set of \( M \) experts.
The router computes a gating score vector as follows:
\begin{equation}
\mathbf{g}(x) = \text{Gate}(x) \in \mathbb{R}^M
\label{eqaul1}
\end{equation}
where \( \mathbf{g}(x) = [g_1(x), g_2(x), \ldots, g_M(x)] \), and \( g_m(x) \) is the importance score assigned to expert \( E_m \). 
These scores are used to determine which subset of experts should be activated to process the input, typically selecting the top-$K$ experts or applying a sparse attention mechanism for computational efficiency.

\subsection{Pretraining and Transfer Learning}
The standard training protocol for foundation models typically consists of two stages: a large-scale pretraining phase and a downstream adaptation phase. During pretraining, the model is trained on broad, unlabeled graph data to learn general-purpose representations. Pretraining stage is formalized as:
\begin{equation}
\min_{\theta} \ \mathcal{L}_{\text{pre}}(\Phi_\theta)
\end{equation}
where \( \Phi_\theta \) denotes the encoder with parameters \( \theta \). The pretraining loss \( \mathcal{L}_{\text{pre}} \), for example, in the case of contrastive pretraining ~\cite{GraphCL}, it can be defined as:
\begin{equation}
    \mathcal{L}_{\mathrm{pre}} = -\log \frac{\exp(\mathrm{sim}(\mathbf{h}_i, \mathbf{h}_i^+))}{\sum_{j} \exp(\mathrm{sim}(\mathbf{h}_i, \mathbf{h}_j))}
\end{equation}
where $\mathbf{h}_i$ and $\mathbf{h}_i^+$ are the representations of positive pairs, and $\mathrm{sim}(\cdot, \cdot)$ denotes a similarity function (e.g., cosine similarity).\\
The downstream adaptation stage is formalized as:
\begin{equation}
\min_{\theta,\phi} \ \mathcal{L}_{\text{task}}(f_\phi(\Phi_\theta))
\end{equation}
where \( f_\phi \) is a task-specific prediction head with parameters \( \phi \), and \( \mathcal{L}_{\text{task}} \) is a supervised loss corresponding to the downstream task. For example, for node or graph classification, $\mathcal{L}_{\mathrm{task}}$ can be instantiated as the cross-entropy loss:
\begin{equation}
\mathcal{L}_{\mathrm{task}} = -\sum_{i} y_i \log \hat{y}_i
\end{equation}
where $y_i$ is the ground-truth label and $\hat{y}_i$ is the predicted probability.
\section{Methodology}

Fig.~\ref{fig2} provides an overview of the proposed GMoPE framework, which unifies graph prompting and MoE to enable scalable and transferable graph representation learning. GMoPE operates in three stages: (a) Pre-training jointly optimizes all experts and their prompts, guided by a structure-aware router and a soft orthogonality loss to enhance expert diversity; (b) Transfer learning freezes expert parameters, fine-tuning only lightweight prompts for efficient adaptation to new tasks; (c) Inference leverages the router to aggregate outputs from prompt-conditioned experts, generating final graph representations for diverse downstream tasks. This design ensures strong compatibility and extensibility across graph domains.\\
\subsection{Feature Alignment}
To enable joint pretraining across multiple graph datasets, feature alignment is essential due to substantial variations in node feature dimensionality and graph size (i.e., number of nodes). We address this by introducing a unified feature alignment layer that projects raw node features from each dataset into a shared latent space. This ensures cross-domain compatibility and facilitates parameter sharing in downstream expert models.\\
Formally, the alignment transforms input features as follows:

\begin{equation}
\tilde{X}^{(i)} = \text{Proj}(X^{(i)}) \in \mathbb{R}^{|\mathcal{V}^{(i)}| \times d_0}
\end{equation}
where \( X^{(i)}  \) is the original node feature matrix of the \( i \)-th graph, and \( \tilde{X}\) is the projected feature matrix with unified embedding dimension \( d_0 \).
The projection function \( \text{Proj}(\cdot) \) denotes a certain projection operation.
 For all experiments, we adopt SVD~\cite{svdalignment} for feature alignment, as it is widely used in graph learning and introduces no additional parameters, thus avoiding extra complexity and instability in MoE-based models.
\subsection{Expert Prompt}
Unlike prior approaches that employ a shared prompt or task-specific prompts for a single GNN backbone, our design dedicates a unique learnable prompt vector to each expert within the MoE architecture. This per-expert prompting strategy achieves two critical objectives: (1) It encourages specialized expertise by directing each expert towards distinct subspaces of the input data, effectively mitigating expert collapse; and (2) It enables expert-specific semantic adaptation, allowing the ensemble to capture the diverse structural and semantic patterns inherent in multi-domain graph data more effectively.\\
A learnable prompt vector \( \mathbf{p}_m \in \mathbb{R}^{d_p} \) is assigned to each expert model \( E_m \). We concatenate the prompt vector \( \mathbf{p}_m \) with the aligned node feature matrix \( \tilde{X}^{(i)} \in \mathbb{R}^{|\mathcal{V}^{(i)}| \times d_p} \) from the \( i \)-th graph. 
The resulting prompt-enhanced input is defined as:
\begin{equation}
\hat{X}^{(i)}_m = [\tilde{X}^{(i)} \, \| \, \mathbf{p}_m \mathbf{1}^\top] \in \mathbb{R}^{|\mathcal{V}^{(i)}| \times (d_0+d_p)}
\end{equation}
where \( \mathbf{1}^\top \in \mathbb{R}^{1 \times |\mathcal{V}^{(i)}|} \) is a row vector of all ones used to broadcast \( \mathbf{p}_m \) across all nodes. This prompt-augmented input \( \hat{X}^{(i)}_m \) is then forwarded to the corresponding expert model \( E_m \) for message passing. \\
During pre-training, prompt vectors are jointly optimized with model parameters to capture transferable, high-level semantic patterns across diverse graph distributions. This enables dynamic expert selection based on input characteristics. For downstream adaptation, only the prompt vectors and task-specific components (e.g., classification heads) require fine-tuning, while core model parameters remain fixed. This parameter-efficient design leverages prompts as lightweight adaptive interfaces that reconfigure model behavior for target tasks, effectively resolving \textbf{C3}.\\
While our prompt structure builds upon the GPF~\cite{GPF}, we extend it by integrating a MoE architecture. ssociating each expert with a unique prompt vector allows GMoPE to model a richer function space than GPF, thereby achieving strictly greater representational expressiveness. \\
\begin{theorem}
Let $\mathcal{F}_{\mathrm{GPF}}$ and $\mathcal{F}_{\mathrm{GMoPE}}$ denote the function classes induced by GPF and GMoPE, respectively. Then the following conclusion holds:
\begin{equation}
    \mathcal{F}_{\mathrm{GPF}} \subsetneq \mathcal{F}_{\mathrm{GMoPE}}
\end{equation}
\end{theorem}
The proof of this theorem can be found in Appendix A.

\subsection{Structure-Aware MoE Routing}
Our framework employs a structure-aware router that dynamically assigns inputs to experts based on graph distributions. This improves representational capacity while avoiding forced generalization across incompatible topologies~\cite{autoMoE}.

Our method directly incorporates expert output into routing scores.
This synergistic design yields semantically coherent expert activation and allows prompt signals to dynamically guide structural specialization (Fig.~\ref{fig2} (a) and (b)), resolving the topology-routing misalignment problem.
\subsubsection{Soft Router}
Firstly, \( \text{Rawscore}_m \) represents the structure-aware rating of the model obtained on a certain batch  \( \mathcal{B} \) and  and $B = |\mathcal{B}|$ is the batch size: :
\begin{equation}
\text{Rawscore}_m = \frac{1}{B} \sum_{s \in \mathcal{B}} \mathcal{L} \left( E_m\left(\hat{X}_m^{(i)}; s \right) \right)
\end{equation}
The loss \( \mathcal{L} \) here can represent the cross entropy loss of GAE~\cite{GAE}, the contrastive loss of GraphCL~\cite{GraphCL} or other strategies. Then we select the top-\(K\) experts with the highest Rawscore, where \(K\) is a hyperparameter ($K < M$).
\begin{equation}
\mathbf{g}(x)_i = 
\begin{cases}
\displaystyle\frac{\exp(\text{Rawscore}_i / \tau)}{\sum_{j \in \mathcal{K}} \exp(\text{Rawscore}_j / \tau)} & \text{if } i \in \mathcal{K} \\
0 & \text{otherwise}
\end{cases}
\end{equation}
where \( \tau \) is a temperature hyperparameter, and \(\mathcal{K}\) denotes the indices set of experts selected by the top-$K$ Rawscore. We preferentially employ this strategy under data-abundant conditions (e.g., during pretraining), where sufficient samples enable finer-grained expert specialization.
\subsubsection{Hard Router}
Similarly, after computing the Rawscore, we select the top-\(K\) experts with the highest values (hard routing). This strategy differs from soft routing and is formally defined as:
\begin{equation}
\mathbf{g}(x)_i = 
\begin{cases}
\displaystyle \frac{1}{K} & \text{if } i \in \mathcal{K} \\
0 & \text{otherwise}
\end{cases}
\end{equation}
This strategy is optimally deployed in low-resource settings (e.g., downstream transfer learning), where it ensures adequate parameter updates despite limited training samples, mitigating optimization instability.\\
\subsection{Embedding Aggregation}
During inference, each expert \( E_i \) processes its prompt-enhanced input and produces an embedding to generate an embedding \( \mathbf{h}_i \in \mathbb{R}^d \). While uniform averaging of expert outputs is common~\cite{nlpmoebasepaper}, this approach risks diluting high-quality predictions from top-performing experts, as it ignores their specialized structural expertise. Similarly, naive voting strategies fail to weight experts by task-specific competence.
To address this, we propose a confidence-guided aggregation strategy leveraging expert predictive certainty.\\
As shown in Fig.\ref{fig2}(c), for expert \( E_m \)'s task prediction \( \hat{\mathbf{y}}_m \), we compute its confidence score as:
\begin{equation}
\begin{aligned}
\alpha_m &= 1 - \mathcal{H}(\hat{\mathbf{y}}_m) \\
\mathcal{H}(\hat{\mathbf{y}}_m) &= -\frac{1}{\log C} \sum_{c=1}^{C} \hat{y}_{m,c} \log \hat{y}_{m,c}
\end{aligned}
\end{equation}
where $\mathcal{H}(\cdot)$ denotes entropy,explicitly quantifying prediction uncertainty. $C$ is the number of classes.
We then compute normalized confidence weights:
\begin{equation}
\begin{aligned}
\omega_i = \frac{\alpha_i}{\sum\limits_{j=1}^{M} \alpha_j}
\end{aligned}
\end{equation}
Finally, the aggregated representation is computed as follows: 
\begin{equation}
\mathbf{h}_{\text{final}} = \sum_{i=1}^M \omega_i \cdot \mathbf{h}_i
\end{equation}
\\
The structure-aware MoE router exhibits inherent compatibility with our per-expert prompting strategy. This synergistic integration leverages both structural topology and semantic cues to guide expert activation: structural signals from the router dynamically select experts based on graph topology;
prompt conditioning adapts expert semantics to task-specific contexts.
The joint optimization of these components enables:
finer-grained specialization through topology-aware semantic adaptation;
Enhanced robustness across diverse graph domains and tasks. This unified framework thereby resolves \textbf{C1}.

\subsection{Overcoming Expert Collapse}
Although load balancing penalties~\cite{AnyGraph} are commonly used to mitigate expert collapse by enforcing uniform routing distributions, this approach ignores potential nonuniformity in underlying data structures and task requirements. Instead, we promote expert specialization through explicit prompt diversification: by maximizing pairwise orthogonality among prompt vectors, we ensure that each expert receives a distinct structural-semantic conditioning signal. This prompt diversity implicitly balances expert utilization: when prompts become highly orthogonal, the router naturally distributes inputs across experts without explicit regularization. Thus, each expert develops domain-specific competence, inherently reducing collapse risk.\\
Prompt effectiveness requires  \( d_p \) (prompt dimension) to substantially exceed the aligned feature dimension. This typically results in   $d_p > M$, where strict orthogonality becomes statistically unviable—high-dimensional random vectors exhibit near-orthogonality by default. We therefore introduce a soft orthogonality loss:
\begin{equation}
\label{equa12}
\mathcal{L}_{\text{ortho}} = \frac{1}{M(M-1)} \sum_{\substack{m, n = 1 \\ m \neq n}}^{M} \exp\left( \frac{\mathbf{p}_m^\top \mathbf{p}_n}{\|\mathbf{p}_m\|_2 \, \|\mathbf{p}_n\|_2} \right)
\end{equation}
By introducing this pseudo-orthogonality constraint, we guarantee structural diversity and expert specialization across experts, thereby substantially alleviating expert collapse. In this way, \textbf{C2} is effectively resolved.
\subsection{Loss Function}
Our model training process is divided into pre-training and transfer learning stages:
In the first stage, we    define the training objectives as follows:
\begin{equation}
\min_{\theta,\, \mathbf{p}} \;\lambda \, \mathcal{L}_{\mathrm{ortho}}+ \frac{1}{BM} \sum_{i=1}^{B} \sum_{m=1}^{M} 
g_m^i(\hat{X}_m^i) \; \mathcal{L}_{\mathrm{pre}}\left(\Phi_{\theta_m}\right) 
\end{equation}
where $\lambda$ is a hyperparameter used to control soft orthogonal loss.\\
To address \textbf{C3}, our framework ensures lightweight adaptation by freezing all expert parameters during transfer learning. Only task-specific prompts  and classification heads are optimized. The training objective is:

\begin{equation}
\min_{\mathbf{p},\,\phi} \;\lambda \, \mathcal{L}_{\mathrm{ortho}}+ \frac{1}{BM} \sum_{i=1}^{B} \sum_{m=1}^{M} 
g_m^i(\hat{X}_m^i) \; \mathcal{L}_{\mathrm{task}}(f_\phi\left(\Phi_{\theta_m}\right)) 
\end{equation}

\subsection{Time complexity analysis}

\subsubsection{Pre-processing}
In both the pre-training and downstream fine-tuning stages, our framework includes a pre-processing step that consists of SVD-based feature alignment and prompt augmentation. The SVD~\cite{svdalignment} operation reduces the input feature dimension from $D$ to $d_0$ with a complexity of $O(|V| \cdot D \cdot d_0)$, and the prompt augmentation increases the feature dimension to $d = d_0 + d_p$ with a complexity of $O(|V| \cdot d_p)$. Where $d_0$ is the target dimension after SVD, which is typically set much smaller than the original node feature dimension in conventional datasets, and $d_p$ is the prompt concatenation dimension, which generally satisfies $d_p < d_0$. As a result, the overall feature dimension $d$ remains small.

As these operations are performed only once before training or fine-tuning, their computational cost is negligible compared to the main training and inference procedures.
\subsubsection{Pre-training stage}
We analyze the per-batch computational complexity using a GAE-style link prediction task as an example. The cost is divided into two parts: forward and backward propagation.

\textbf{Forward propagation:} All $M$ experts process the sampled subgraph(denoted as $(\mathcal{V}_\mathcal{B}, \mathcal{E}_\mathcal{B})$) to compute their individual losses, with a total complexity of
\[
O\left(M \cdot \left[L|\mathcal{V}_\mathcal{B}|d^2 + L|\mathcal{E}_\mathcal{B}|d\right]\right)
\]
where $L$ denotes the number of layers in the model. 

\textbf{Backward propagation:} Only the top-$K$ experts are selected for backpropagation and parameter updates, resulting in a reduced complexity of
\[
O\left(K \cdot \left[L|\mathcal{V}_\mathcal{B}|d^2 + L|\mathcal{E}_\mathcal{B}|d\right]\right)
\]

\textbf{Total complexity per batch:}
\[
O\left((M + K) \cdot \left[L|\mathcal{V}_\mathcal{B}|d^2 + L|\mathcal{E}_\mathcal{B}|d\right]\right)
\]

\subsubsection{Downstream Fine-tuning stage}
During downstream task adaptation, GMoPE freezes all GNN expert parameters and only fine-tunes the prompt vectors and the task-specific classification head. Assuming each expert is a 3-layer GCN, the trainable parameters during fine-tuning mainly come from the prompt vectors (typically of dimension $d_p$) and the classification head, while all expert parameters remain fixed.

Empirically, we set the prompt dimension $d_p$ to be between $\frac{1}{4}$ and $\frac{1}{2}$ of the node feature dimension $d_0$ to achieve optimal performance. Under this configuration, the number of trainable parameters during fine-tuning is typically less than $1\%$ of that required for full-parameter fine-tuning. This proportion becomes even smaller when more complex expert models are used, further demonstrating the efficiency and modularity of our framework.

\section{Experiment}

\begin{table*}[ht]
\centering
\begin{tabular}{llcccccc}
\specialrule{1.5pt}{0pt}{0pt}
\makecell{Pre-training\\Strategy} & \makecell{Tuning\\Strategy} & \makecell{Cora} & \makecell{Citeseer} & \makecell{Pubmed} & \makecell{Photo} & \makecell{Computers}  & \makecell{AVG} \\
\midrule

\multirow{5}{*}{DGI}
& FT           & \underline{87.51} & \textbf{87.67} & \textbf{90.87} & \textbf{87.82} & 86.31 & \underline{88.04} \\
& GraphPrompt  & 82.02 & 84.17 & 87.79 & 84.16 & 85.36 & 84.70 \\
& GPF          & 84.93 & 86.57 & \underline{89.87} & 86.02 & 85.74 & 86.63 \\
& AnyGraph     & 84.39 & 83.39 & 87.85 & 85.32 & \underline{86.37} & 85.46 \\
& GMoPE         & \textbf{89.10} & \underline{86.64} & 89.30 & \underline{87.50} & \textbf{88.57} & \textbf{88.22} \\

\midrule

\multirow{5}{*}{GAE}
& FT           & 86.71 & 86.98 & 90.32 & \underline{92.21} & \underline{90.51} & 89.35 \\
& GraphPrompt  & 87.65 & 85.65 & 87.01 & 82.35 &  84.21 & 85.37     \\
& GPF          & \underline{88.20} & \underline{87.56} & 91.02 & 91.34 & 89.77 & \underline{89.58} \\
& AnyGraph     & 84.49 & 86.54 & \underline{91.74} & 90.70 & 89.46 & 88.59 \\
& GMoPE         & \textbf{89.55} & \textbf{88.97} & \textbf{92.67} & \textbf{94.42} & \textbf{93.41} & \textbf{91.80} \\

\specialrule{1.5pt}{0pt}{0pt}

\end{tabular}

\caption{AUC of link prediction task. The first and second highest scores are represented by \textbf{bold font} and \underline{underline}, respectively. AVG denotes the average value.}
\label{linkpred}
\end{table*}

\begin{table*}[ht]
\centering
\resizebox{\textwidth}{!}{
\begin{tabular}{llcccccc|cccc}
\specialrule{1.5pt}{0pt}{0pt}

\makecell{Pre-training\\Strategy} & \makecell{Tuning\\Strategy} &{Cora} &{Citeseer} &{Pubmed} &{Photo} &{Computers} &{AVG} & {PROTEINS} & {DD} & {NCI109} & {AVG} \\

\midrule

\multirow{5}{*}{DGI/GraphCL} 
& FT           &\textbf{73.18} &56.54 &\textbf{70.61} &62.29 &50.73 &62.67 &68.47 &\textbf{65.54} &\textbf{63.76} &\underline{65.92}\\
& GraphPrompt  & 56.32 & 48.01 & 51.86 & 65.86 & 44.49 & 53.31 &  68.45 &  60.37 & 61.68 & 63.50  \\
& GPF          &65.32 &\underline{62.46} &69.62 &\underline{66.01} &\textbf{51.67} &\underline{63.02} &\underline{71.32} &61.11 &62.42 &64.95 \\
& AnyGraph     &63.84 &57.06 &68.17 &63.96 &50.91 &60.79 &71.17 &60.68 &60.61 &64.15 \\
& GMoPE         &\underline{69.37} &\textbf{62.86} &\underline{70.12} &\textbf{69.61} &\underline{50.91} &\textbf{64.57} &\textbf{72.18} &\underline{65.39} &\underline{62.58} &\textbf{66.72} \\

\midrule

\multirow{5}{*}{EdgePred} 
& FT           &\underline{72.69} &58.35 &\textbf{73.46} &\textbf{75.10} &\underline{65.75} &\underline{69.07} &70.72 &\textbf{64.48} &60.80 &\underline{65.33} \\
& GraphPrompt  & 59.84 & 54.69 & 56.39 & 56.83 & 64.32 & 58.41 & 67.59 & 57.64 & 59.32 &  61.52    \\
& GPF          &60.70 &57.14 &69.87 &68.56 &63.20 &63.89 &\underline{71.37} &57.27 &60.86 &63.17 \\
& AnyGraph     &62.98 &\underline{58.92} &70.63 &70.87 &64.38 &65.56 &69.24 &58.82 &\underline{61.17} &63.08 \\
& GMoPE         &\textbf{73.22} &\textbf{62.31} &\underline{71.41} &\underline{73.20} &\textbf{66.11} &\textbf{69.25} &\textbf{76.38} &\underline{61.54} &\textbf{62.53} &\textbf{66.82} \\

\specialrule{1.5pt}{0pt}{0pt}
\end{tabular}
}
\caption{Accuracy (ACC) of classification tasks. DGI and GraphCL pre-training  is used for node  and graph classification task, respectively. EdgePred pre-training is included for both node and graph classification tasks.}
\label{classify}
\end{table*}

\begin{table}[ht]
\centering
\scriptsize
\begin{tabular}{lcccccc}
\specialrule{1.5pt}{0pt}{0pt}
& \multicolumn{2}{c}{Link Pred.} & \multicolumn{2}{c}{Node Cls.} & \multicolumn{2}{c}{Graph Cls.} \\
\cmidrule(lr){2-3} \cmidrule(lr){4-5} \cmidrule(lr){6-7}
Method & DGI & GAE & DGI & EdgePred & GraphCL & EdgePred \\
\midrule
w/ Prompt   & \textbf{88.22} & \textbf{91.80} & \textbf{64.57} & \textbf{69.25} & \textbf{66.72} & \textbf{66.82} \\
w/o Prompt  & 79.96 & 82.96 & 58.41 & 63.95 & 58.19 & 60.85 \\
\specialrule{1.5pt}{0pt}{0pt}
\end{tabular}
\caption{Ablation study on expert prompts across different tasks.}
\label{promptablation}
\end{table}

To evaluate downstream transfer effectiveness, we benchmark our framework on three fundamental graph learning tasks:
link prediction, node classification, and graph classification.
\subsection{Datasets}
To systematically evaluate cross-domain capability, we benchmark our architecture on graph datasets spanning three distinct fields: citation networks, commodity networks, and biomolecule networks.

\subsection{Benchmarks}
To demonstrate the advantages of our approach over existing prompt tuning and adaptation strategies, we conduct comprehensive evaluations along two key axes: Pretraining Strategies and Adaptation Techniques. The first axis encompasses various pretraining paradigms, including GAE~\cite{GAE}, DGI~\cite{DGI},  GraphCL~\cite{GraphCL} and EdgePred~\cite{a2}. 
The latter includes full parameter Fine-tuning (FT)\cite{FT}, GraphPrompt\cite{graphprompt}, GPF~\cite{GPF}, and AnyGraph~\cite{AnyGraph}.

\subsection{Hyperparameter Setting}
As detailed above, our architecture supports flexible instantiation of experts using any GNN variant. To ensure experimental consistency and fair comparison, all models (including baselines) implement experts as GCN~\cite{a2}.

Hyperparameters are systematically optimized for every method through grid search to achieve the best performance. For our approach, we implement dynamic adjustment of the regularization coefficient \(\lambda\),  number of experts, batch size, and temperature parameter. All tabulated results also present means computed over 5  independent runs with different random seeds.
\subsection{Result Analysis}
\subsubsection{Link Prediction}
As shown in Table~\ref{linkpred}, GMoPE  achieves promising results on link prediction task across all datasets and pretraining strategies. GMoPE  considerably outperforms both GPF and GraphPromt in most cases. On average, GMoPE even surpasses full parameter fine-tuning. Compared to the recent AnyGraph, GMoPE achieves a 3.42\% performance gain. These results validate GMoPE's superior cross-domain generalization and lightweight adaptation capability for edge-level tasks.
\subsubsection{Classification}
As shown in Table~\ref{classify}, GMoPE consistently achieves either the best or highly competitive performance across both node-level and graph-level classification tasks under all pretraining strategies. For instance, GMoPE achieves up to 19.84\% and 6.84\% improvement over GraphPromt in node and graph classification task, respectively. These results highlight GMoPE’s dual strength: robust generalization across different task granularities and superior adaptation capabilities.
\subsection{Ablation Analysis}
\subsubsection{MoE framework}
When $M=1$ (single expert), our MoE framework reduces to the GPF baseline. Our consistent out performance of GPF across all cross-domain tasks demonstrates that the effectiveness of adopting the MoE paradigm, as the incorporation of multiple experts significantly enhances the model’s generalization ability in cross-domain scenarios.
\subsubsection{Expert prompt}
Table~\ref{promptablation} compares our model’s performance with and without expert prompts across link prediction, node classification, and graph classification tasks. Removing expert prompts (\textit{w/o Prompt}) prevents prompt-based parameter-efficient adaptation, requiring all expert parameters to be fine-tuned for each task. This leads to clear performance degradation and increased computational overhead, primarily due to the loss of expert specialization and less effective knowledge transfer. In contrast, incorporating expert prompts (\textit{w/ Prompt}) maintains expert diversity and enables efficient adaptation by only tuning prompts and task heads. These results underscore the importance of expert prompts for  the effectiveness of MoE-based models.

\begin{figure}[t]
\centering
\includegraphics[width=0.8\columnwidth]{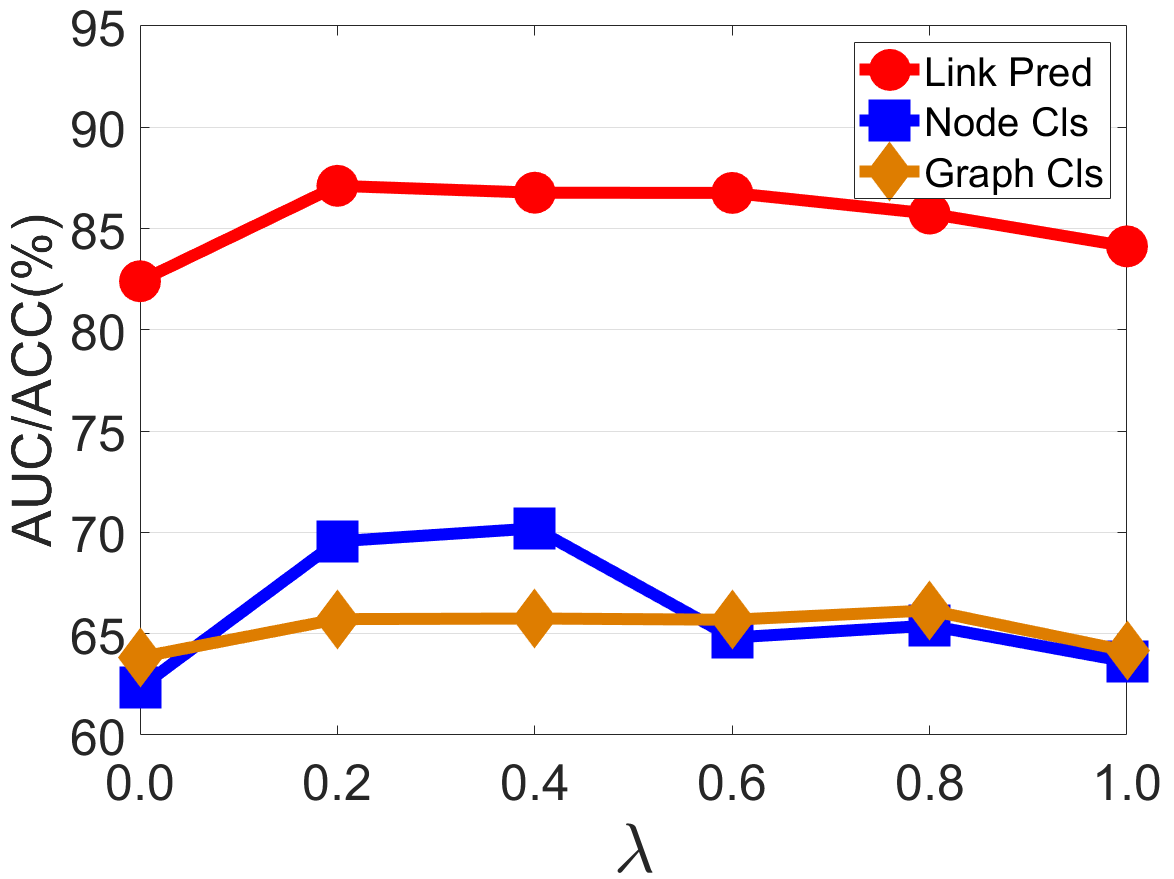}
\caption{The performance  impact of soft orthogonal loss on different downstream tasks.}
\label{fig3}
\end{figure}


\subsubsection{Soft orthogonal loss}
We analyze the impact of the orthogonality loss coefficient 
\(\lambda\) (Fig. \ref{fig3}). 
 When 
\(\lambda\)  is too small, expert specialization becomes insufficient, which hinders the MoE’s ability to leverage diverse expertise.
In contrast, when \(\lambda\) is excessively large, it leads to unstable prompt vector oscillations, causing erroneous feature enhancement and performance degradation.  \\

\section{Conclusion}
We propose GMoPE, a graph foundation model unifying MoE with prompt tuning. Key innovations include: Expert-specific prompts enabling specialized subdomain adaptation;
Structure-aware routing for dynamic expert selection; 
Soft orthogonality loss preventing expert collapse.
Experiments show state-of-the-art generalization and transfer efficiency versus prompt/MoE baselines. Ablations confirm criticality of all components.
GMoPE delivers scalable universal graph learning, with future work on dynamic architectures and heterogeneous/dynamic graphs.


\bibliography{aaai2026}

\clearpage
\appendix

\section{A.Theoretical Analysis}
\subsection{Theorem Proof}

Here, we provide a detailed proof and explanation for the Theorem 1.\\
\textbf{Lemma 1 Expressiveness of Concatenation vs. Addition:} \\ 
Let $x \in \mathbb{R}^{d_0}$ be a node feature and $p \in \mathbb{R}^{d_p}$ a prompt vector. The concatenation operation $[x; p]$ is strictly more expressive than the addition operation $x + p'$ (where $p' \in \mathbb{R}^{d_0}$), in the sense that any function of the form $f(x + p')$ can be represented as $g([x; p])$ for some $g$, but not vice versa~\cite{promptpower}.\\
\textbf{Lemma 2 (Expressiveness of MoE Routing):}
Let $\mathcal{F}_{\mathrm{single}}$ denote the function class realizable by a single expert (e.g., a single MLP or GNN), and let $\mathcal{F}_{\mathrm{MoE}}$ denote the function class realizable by a Mixture-of-Experts (MoE) model with $M$ experts and a routing function. The MoE model computes
\begin{equation}
f_{\mathrm{MoE}}(x) = \sum_{j=1}^M \alpha_j(x) f_j(x),
\end{equation}
where $f_j \in \mathcal{F}_{\mathrm{single}}$, $\alpha_j(x) \geq 0$, and $\sum{j=1}^M \alpha_j(x) = 1$ for all $x$, with $\alpha_j(x)$ determined by the routing function (which may be hard or soft). Then,
\begin{equation}
\mathcal{F}_{\mathrm{single}} \subsetneq \mathcal{F}_{\mathrm{MoE}}.
\end{equation}
\textit{Proof:}
It is straightforward to see that $\mathcal{F}_{\mathrm{single}} \subseteq \mathcal{F}_{\mathrm{MoE}}$, since for any $f \in \mathcal{F}_{\mathrm{single}}$, we can construct an MoE function by setting $f_1 = f$, $f_j = 0$ for $j \neq 1$, and $\alpha_1(x) = 1$, $\alpha_j(x) = 0$ for $j \neq 1$, so that $f_{\mathrm{MoE}}(x) = f(x)$.\\
To show that the inclusion is strict,we exhibit a function $f^* \in \mathcal{F}_{\mathrm{MoE}}$ that cannot be represented by any single expert in $\mathcal{F}_{\mathrm{single}}$.\\
Consider the case where $M=2$, and let $f_1, f_2 \in \mathcal{F}_{\mathrm{single}}$ be two distinct functions ($\exists x$ let $f_1(x) \neq f_2(x)$). Define a routing function such that for some non-empty subset $\mathcal{X}_1 \subset \mathcal{X}$ (where $\mathcal{X}$ is the input space), $\alpha_1(x) = 1$ and $\alpha_2(x) = 0$ for all $x \in \mathcal{X}_1$, and $\alpha_1(x) = 0$ and $\alpha_2(x) = 1$ for all $x \in \mathcal{X} \setminus \mathcal{X}_1$. Then, the MoE function is:
\begin{equation}
f_{\mathrm{MoE}}(x) = \begin{cases}
f_1(x) & \text{if } x \in \mathcal{X}_1, \\
f_2(x) & \text{otherwise.}
\end{cases}
\end{equation}
Assume for contradiction that some $f \in \mathcal{F}_{\mathrm{single}}$ satisfies $f(x) = f_{\mathrm{MoE}}(x)$ for all $x \in \mathcal{X}$. This would require $f$ to simultaneously match $f_1$ on $\mathcal{X}_1$ and $f_2$ elsewhere, which is impossible when $\mathcal{F}_{\mathrm{single}}$ is constrained.\\
\textbf{By Lemma 1 and Lemma 2:}  
Any function that can be represented by GPF (i.e., addition-based prompt with a single expert) can also be represented by GMoPE (i.e., concatenation-based prompt with MoE), by simply using one expert and setting the prompt accordingly. Therefore,
\begin{equation}
    \mathcal{F}_{\mathrm{GPF}} \subseteq \mathcal{F}_{\mathrm{GMoPE}}.
\end{equation}
\textbf{A concrete example:}  
Suppose we have two input domains $\mathcal{D}_1$ and $\mathcal{D}_2$ (e.g., graphs from two different distributions). Assume that that the optimal transformation for $\mathcal{D}_1$ is $g_1([x; p_1]) = x \odot p_1$ and for $\mathcal{D}_2$ is $g_2([x; p_2]) = x + p_2$. The GMoPE model, with an appropriate routing function, can assign inputs from $\mathcal{D}_1$ to expert 1 and from $\mathcal{D}_2$ to expert 2, each with its own prompt and GNN parameters. Thus, the overall function is:
\begin{equation}
    f_{\mathrm{GMoPE}}(x) = \alpha_1(x) g_1([x; p_1]) + \alpha_2(x) g_2([x; p_2]).
\end{equation}

However, GPF is limited to applying a single global transformation of the form $f(x + p)$ across all inputs, and thus cannot model domain-specific, piecewise, or convex-combination behaviors. Consequently, the function in question lies within the function space $\mathcal{F}_{\mathrm{GMoPE}}$ but not within $\mathcal{F}_{\mathrm{GPF}}$.\\
GMoPE activation enables the model to perform group-wise, domain-wise, or even sample-wise feature enhancement, in contrast to the global and uniform transformation applied by GPF. This conditional and adaptive mechanism grants GMoPE fundamentally greater expressive power.
Therefore, we obtain the following strict inclusion:
\begin{equation}
    \mathcal{F}_{\mathrm{GPF}} \subsetneq \mathcal{F}_{\mathrm{GMoPE}}.
\tag{8}
\end{equation}

\section{B.Experimental Details}
\subsection{Datasets}
We provide an explanation of datasets, where their statistics details are presented in Table ~\ref{datasets}.

\begin{table*}[!htbp]
\centering
\footnotesize 
\begin{tabular}{lcccccc} 
\toprule 
Dataset & \#Graphs & Graph classes & Avg. nodes & Avg. edges & Features & Classes \\
\midrule 
Cora         & 1        & -             & 2,708      & 5,429      & 1,433    & 7       \\
Citeseer     & 1        & -             & 3,327      & 4,732      & 3,703    & 6       \\
Pubmed       & 1        & -             & 19,717     & 44,338     & 500      & 3       \\
Photo        & 1        & -             & 7,650      & 119,081    & 745      & 8       \\
Computers    & 1        & -             & 13,752     & 245,778    & 767      & 10      \\
Proteins     & 1,113    & 2             & 39.1       & 72.8       & 1        & -       \\
DD           & 1,178    & 2             & 284.3      & 715.7      & 89       & -       \\
NCI109       & 4,127    & 2             & 29.9       & 32.3       & 38       & -       \\
\bottomrule 
\end{tabular}
\caption{Summary of datasets.}
\label{datasets}
\end{table*}

\subsubsection{Citation network}
Cora and Citeseer \cite{Cora&Citeseer} jointly provide two small-scale corpora of computer-science literature, 
where each node is encoded by a sparse bag-of-words vector and labeled with a corresponding research topic. 
Pubmed \cite{Pubmed} compiles diabetes-focused abstracts from PubMed; each node is represented by a denser TF-IDF vector and assigned one of three diabetes-type labels.
\subsubsection{E-commerce co-purchase graphs}
Computers and Photo are two co-purchase graphs extracted from the Amazon review corpus introduced by \cite{AmazonData}. Each node represents a product, and edges indicate that two products are frequently bought together. Node features are TF-IDF vectors derived from customer reviews, and the task is to classify products into sub-categories.
\subsubsection{Molecular graphs}
PROTEINS\cite{Protein}, DD\cite{DD} and NCI109\cite{NCI109} are collections of molecular or protein graphs. Each node represents an atom, and each edge represents a chemical bond. The task is binary graph-level classification: PROTEINS distinguishes enzymes from non-enzymes, DD separates drug-like molecules from decoys, and NCI109 predicts anti-cancer activity.

\begin{table}[t]
\centering
\begin{tabular}{lcc}
\specialrule{1.5pt}{0pt}{0pt}
\makecell{Tunning\\Strategy} &{Cora} &{Photo} \\
\midrule

FT                &73.62      &62.29        \\
GraphPrompt       &71.37      &59.08             \\
GPF               &73.62      &\underline{68.17}        \\
AnyGraph          &\underline{74.28}      &59.28        \\
GMoPE              &\textbf{75.28}      &\textbf{73.14}        \\

\specialrule{1.5pt}{0pt}{0pt} 
\end{tabular}
\caption{The accuracies achieved after joint pre-training on the Cora and Photo datasets using the DGI strategy.}
\label{simple experiment}
\end{table}

\subsection{Experimental group setting}
It is important to emphasize that our experimental protocol adopts three distinct multi-dataset settings: one group comprising all citation networks, another group for all product networks, and a third group for all graph classification datasets.
This design choice is motivated by the observation that existing SVD-based alignment methods~\cite{svdalignment} lack theoretical guarantees for effective alignment across heterogeneous application domains. In particular, the semantic information encoded in node features varies substantially between these domains. Additionally, the node feature dimensionalities often differ significantly between datasets used for node classification (e.g., citation and product networks) and those used for graph classification, further complicating alignment. \\
To further illustrate the structure-aware capability of our MoE routing, we conducted a preliminary experiment by jointly pre-training on the Cora and Photo datasets—two sources that are, in fact, significantly heterogeneous in terms of their graph structures. The motivation behind this experiment is to demonstrate that our model, equipped with structure-aware MoE routing, can effectively identify and adapt to the substantial structural differences between the two graphs. During training, the routing mechanism enables samples from Cora to be primarily directed to a subset of experts, while samples from Photo are routed to another subset. This sample-wise expert allocation allows the model to handle the heterogeneity between the datasets, leading to strong performance on both domains.\\
The results, presented in Table~\ref{simple experiment}, are based on the DGI pretraining strategy as a representative example. As shown, other methods tend to exhibit a performance bias, typically excelling on Cora while underperforming on Photo. In contrast, our approach achieves consistently strong results on both datasets, even surpassing the results reported earlier in Table~\ref{classify}. This improvement can be reasonably attributed to the structure-aware routing, rather than to SVD-based alignment, as SVD-based methods are not designed to handle such significant cross-domain structural differences. Overall, these findings highlight the superior structure-awareness of our MoE architecture, which enables effective expert specialization and robust performance.\\
In our experiments, we uniformly aligned the feature dimensions across datasets within each domain: citation networks to 196, product networks to 256, and graph classification datasets to 16.
\subsection{Comparison of Parameter Quantities}
We compared the total number of learnable parameters between our method and several baselines during the tuning phase of three downstream tasks.\\
As shown in Table~\ref{Number of Parameters}, the number of learnable parameters in our approach is of the same order of magnitude as those in the two traditional prompt-based methods—GPF and GraphPrompt. Despite this, our method achieves superior performance, as demonstrated in Tables~\ref{linkpred} and~\ref{classify}.
Furthermore, our parameter count is substantially lower than that of FT and AnyGraph. In particular, compared to FT, the number of parameters tuned in our method is less than 1\% of that in FT.\\
Importantly, the number of learnable parameters in our method does not scale explosively with either the dataset size or the dimensionality of node features, highlighting its scalability and efficiency.

\begin{table}[t]
\centering
\small
\begin{tabular}{@{}lccc@{}}
\specialrule{1.5pt}{0pt}{0pt}
Model & Link Pred & Node Cls & Graph Cls \\
\midrule
FT        & 20,928 & 20,928 & 3,584 \\
GraphPrompt        & 32     & 32     & 32 \\
GPF       & 196    & 196    & 16 \\
AnyGraph  & 62,784 & 62,784 & 10,752 \\
GMoPE     & 192    & 192    & 12 \\
\specialrule{1.5pt}{0pt}{0pt}
\end{tabular}
\caption{The number of learnable parameters in downstream task adaptation. Node classification and link prediction are based on experimental settings on citation networks, while graph classification corresponds to experiments on all graph classification datasets.}
\label{Number of Parameters}
\end{table}

\subsection{Details of Pretraining Methods}
\begin{itemize}
\item \textbf{GAE}~\cite{GAE} The unsupervised autoencoder model learns node representations that capture and reconstruct the graph's structural and semantic information.
\item \textbf{DGI}~\cite{DGI} DGI maximizes mutual information between node representations and the global graph summary by training a discriminator to distinguish between real and corrupted graph samples.
\item \textbf{GraphCL}~\cite{GraphCL} A contrastive self-supervised framework generates augmented graph views via random perturbations and trains a graph neural network to maximize agreement between representations of the same graph while minimizing agreement with others.
\item \textbf{EdgePred}~\cite{a2} Supervised link prediction is commonly used in GNN pre-training due to abundant labels, and its results on downstream tasks help assess cross-task transfer learning performance.
\end{itemize} 
All experiments were carried out on two separate machines: (1) an Intel® CoreTM i9-12900K CPU with dual NVIDIA GeForce RTX 3090 GPUs (24GB VRAM) and 128GB RAM, and (2) an Intel® Xeon® Gold 5218 CPU with an NVIDIA GeForce RTX 4090 GPU (24GB VRAM) and 256GB RAM. The software stack consistently employed PyTorch 1.13.1 and PyTorch Geometric 2.6.1 on both systems to ensure reproducible results. Batch sizes were dynamically adjusted according to each GPU's memory capacity.\\
\begin{figure}[t]
\centering
\includegraphics[width=1\columnwidth]{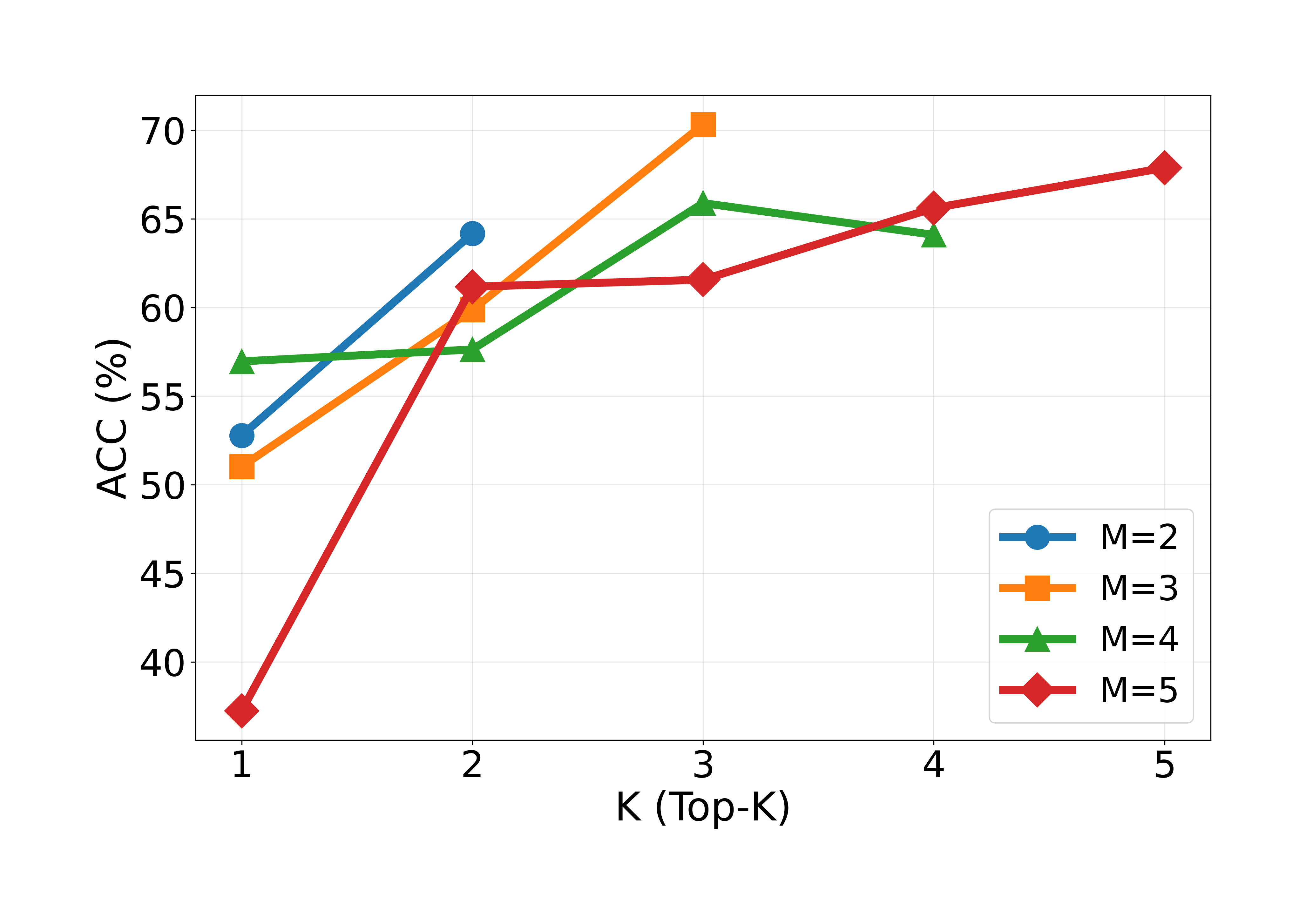}
\caption{Node classification performance under different values of M and K (on citation networks,using DGI pre training strategy)}
\label{MandK}
\end{figure}
\subsection{Hyperparameter Setting Details}

We explain the rationality of different hyperparameter settings and provide appropriate numerical ranges to facilitate the reproduction of our work.\\
In our experiments, we utilize random seeds in the range of 41 to 45. The dataset is randomly partitioned into training, test, and validation sets with a ratio of 7:2:1. During the fine-tuning stage for downstream tasks, a small subset of samples is randomly selected from the training set in each episode. Model selection is conducted based on the performance on the validation set, and the parameters corresponding to the best validation result are retained. Subsequently, the model with the optimal parameters, as determined on the validation set, is evaluated on the test set.
\subsubsection{Number of experts and Top\_K}
For our GMoPE framework, a straightforward approach is to set $M = N$, where $N$ is the number of datasets. This is a natural choice, as inter-dataset differences are typically more significant than intra-dataset variations. If each expert can potentially specialize in a single dataset, it is expected to yield strong performance. \\
The choice of 
$K$, the number of activated experts, is task-dependent. For fine-grained tasks, such as edge-level and node-level prediction, we recommend setting $K=M$ during both the pre-training and downstream fine-tuning stages. This configuration ensures that all experts are sufficiently trained, promoting robust representation learning.
In contrast, for coarser-grained graph-level tasks, we suggest using 
$K=1$ to encourage strong specialization among experts, thereby enhancing differentiation and improving overall performance.\\
It is also important to note that 
$K$ is closely tied to computational efficiency. As 
$K$ increases, the time cost in both the pre-training and downstream adaptation stages grows significantly. Therefore, 
$K$ represents a non-trivial hyperparameter that must balance performance and efficiency.
We recommend beginning with two baseline settings:
$K=1$ and 
$K=M$ to assess initial performance, followed by further tuning based on task-specific requirements and available computational resources.\\
On citation network datasets($N=3$), we conducted performance evaluations for node classification downstream tasks under different settings of $M$ and $K$. The results, as shown in the figure~\ref{MandK}, can serve as a reference for selecting appropriate values of $M$ and $K$. It is evident that when the number of experts $M$ is too small, the potential benefits of the MoE architecture cannot be fully realized. Conversely, when $M$ is too large, some experts inevitably experience insufficient training, which in turn results in suboptimal overall model performance.
\begin{figure}[t]
\centering
\includegraphics[width=0.8\columnwidth]{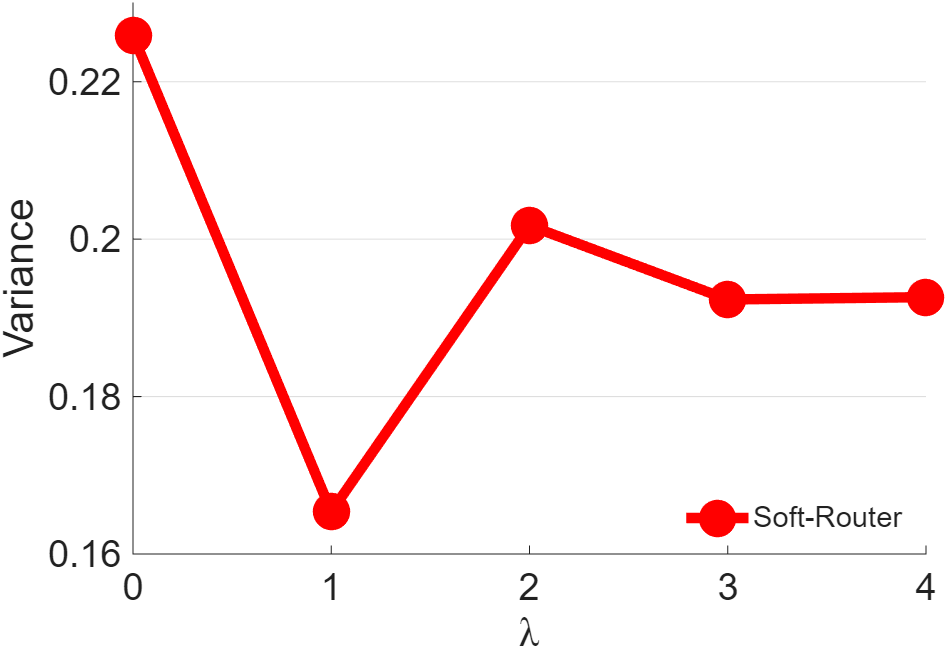}
\caption{The relationship between soft orthogonal loss and soft router weight allocation}
\label{ortho_soft}
\end{figure}

\begin{figure}[t]
\centering
\includegraphics[width=0.8\columnwidth]{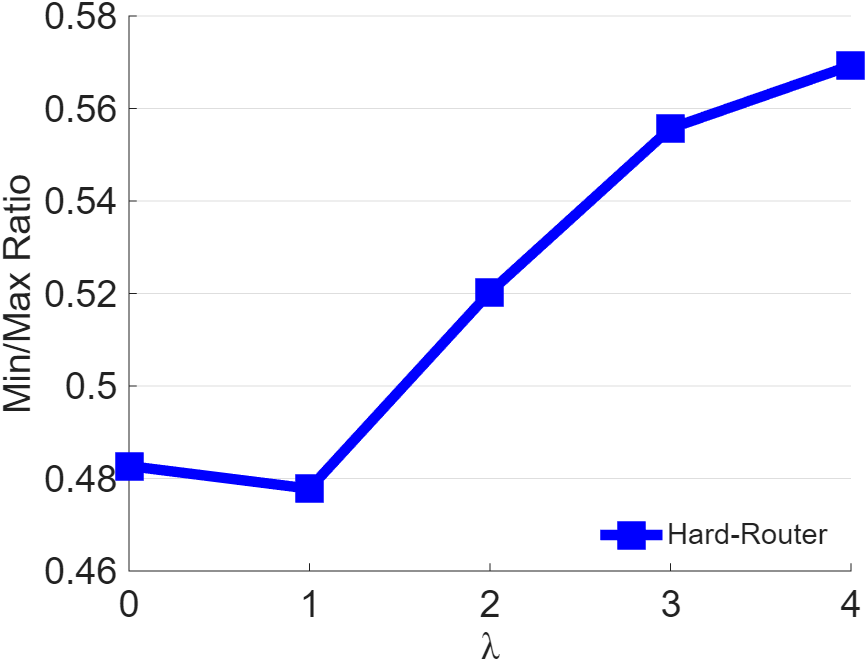}
\caption{The relationship between soft orthogonal loss and hard router weight allocation}
\label{ortho_hard}
\end{figure}
\subsubsection{soft orthogonality loss}
To further investigate the effectiveness of the soft orthogonality loss in addressing ~\textbf{C2} , we analyze the relationship between the loss weight \(\lambda\) and weight allocation with experts.We take k=1, M=2, conducting investigations on citation network datasets using the DGI pre-training strategy, with node classification as the downstream task.The results are shown in Figures ~\ref{ortho_soft} and ~\ref{ortho_hard}.\\
In the pre training stage, we use the variance of expert average weights to evaluate (the closer it is to 0, the more evenly distributed it is).In the downstream stage, we measure it by the ratio of the minimum to maximum number of times experts are selected (the closer it is to 1, the more evenly distributed the allocation).\\
It is evident that, regardless of whether a soft or hard router is used, the soft orthogonality loss plays a notable role in expert weight allocation. This indicates that the soft orthogonality loss can effectively regulate the distribution of expert weights by dynamically controlling the degree of expert specialization.\\
It should be noted, however, that a larger value of $\lambda$ is not always better. Even small changes in $\lambda$ can lead to significant variations in model performance on downstream tasks, as shown in Figure~3. Although adjust $\lambda$ can make the distribution of expert weights more uniform, ``uniformity implies mediocrity''; maintaining a healthy level of competition among experts is also important for achieving strong performance in MoE architectures. Based on our experiments, we find that setting $\lambda$ within the range of $(0, 3)$ generally yields optimal performance across different tasks.

\subsubsection{Other hyperparameters}
There exist several additional hyperparameters, though none exert a decisive influence on model performance. In our experiments, these were held constant. The temperature parameter $\tau$ can marginally enhance expert diversity, yet its impact is substantially weaker compared to $\lambda$; we thus fixed $\tau = 0.8$. The prompt dimension $d_p$ is comparatively more critical. However, due to its intricate functional coupling with our framework, identifying its optimal value is non-trivial. Empirical results suggest that setting $d_p \in \left[\frac{1}{4}d_0, \frac{1}{2}d_0\right]$ delivers satisfactory performance. During the downstream fine-tuning stage, following common practice in previous studies, we set the amount of labeled data to 10\%--20\% of the total size across multiple datasets, corresponding to 200--1000 episodes depending on the datasets.Since the transfer performance of our method on downstream tasks is inherently dependent on the quality of pre-training, we empirically set the number of pre-training epochs within the range of 150–200 to ensure sufficient model convergence while avoiding overfitting.

\end{document}